\documentclass[10pt,twocolumn,letterpaper]{article}

\usepackage{cvpr}              
\usepackage{multirow}
\definecolor{cvprblue}{rgb}{0.21,0.49,0.74}
\usepackage[pagebackref,breaklinks,colorlinks,allcolors=cvprblue]{hyperref}

\def\paperID{13550} 
\def\confName{CVPR}
\def\confYear{2026}

\title{Improving Partially Observed Trajectories Forecasting by Target-driven Self-Distillation}

\author{Peng Shu\qquad Pengfei Zhu\qquad Mengshi Qi\thanks{Corresponding author}\qquad Liang Liu \vspace{1mm} \\
State Key Laboratory of Networking and Switching Technology,\\
Beijing University of Posts and Telecommunications,China\\
{\it \{shup, zhupengfei2000, qms, liangliu\}@bupt.edu.cn}
}

\begin{document}
\maketitle
\begin{abstract}
Accurate prediction of future trajectories of traffic agents is essential for ensuring safe autonomous driving. However, partially observed trajectories can significantly degrade the performance of even state-of-the-art models. Previous approaches often rely on knowledge distillation to transfer features from fully observed trajectories to partially observed ones. This involves firstly training a fully observed model and then using a distillation process to create the final model. While effective, they require multi-stage training, making the training process very expensive. Moreover, knowledge distillation can lead to a performance degradation of the model. In this paper, we introduce a \textbf{T}arget-driven \textbf{S}elf-\textbf{D}istillation method (\textbf{TSD}) for motion forecasting. Our method leverages predicted accurate targets to guide the model in making predictions under partial observation conditions. By employing self-distillation, the model learns from the feature distributions of both fully observed and partially observed trajectories during a single end-to-end training process. This enhances the model's ability to predict motion accurately in both fully observed and partially observed scenarios. We evaluate our method on multiple datasets and state-of-the-art motion forecasting models. Extensive experimental results demonstrate that our approach achieves significant performance improvements in both settings. To facilitate further research, we will release our code and model checkpoints.
\end{abstract}    

\section{Introduction}

For the autonomous driving system, predicting the trajectory of surrounding agents within the next few seconds is a crucial and challenging task. Accurate motion forecasting enhances safety by enabling precise path planning and guiding agents along safe routes, thereby strongly supporting ego-vehicle trajectory planning within the driving pipeline~\cite{jia2023think,hu2023planning,wu2022trajectory}, as well as aiding in robot navigation~\cite{huang2023visual,chen2019crowd} and tracking~\cite{cui2022mixformer}. 

\begin{figure}[!htbp]\centering
\includegraphics[width=0.48\textwidth]{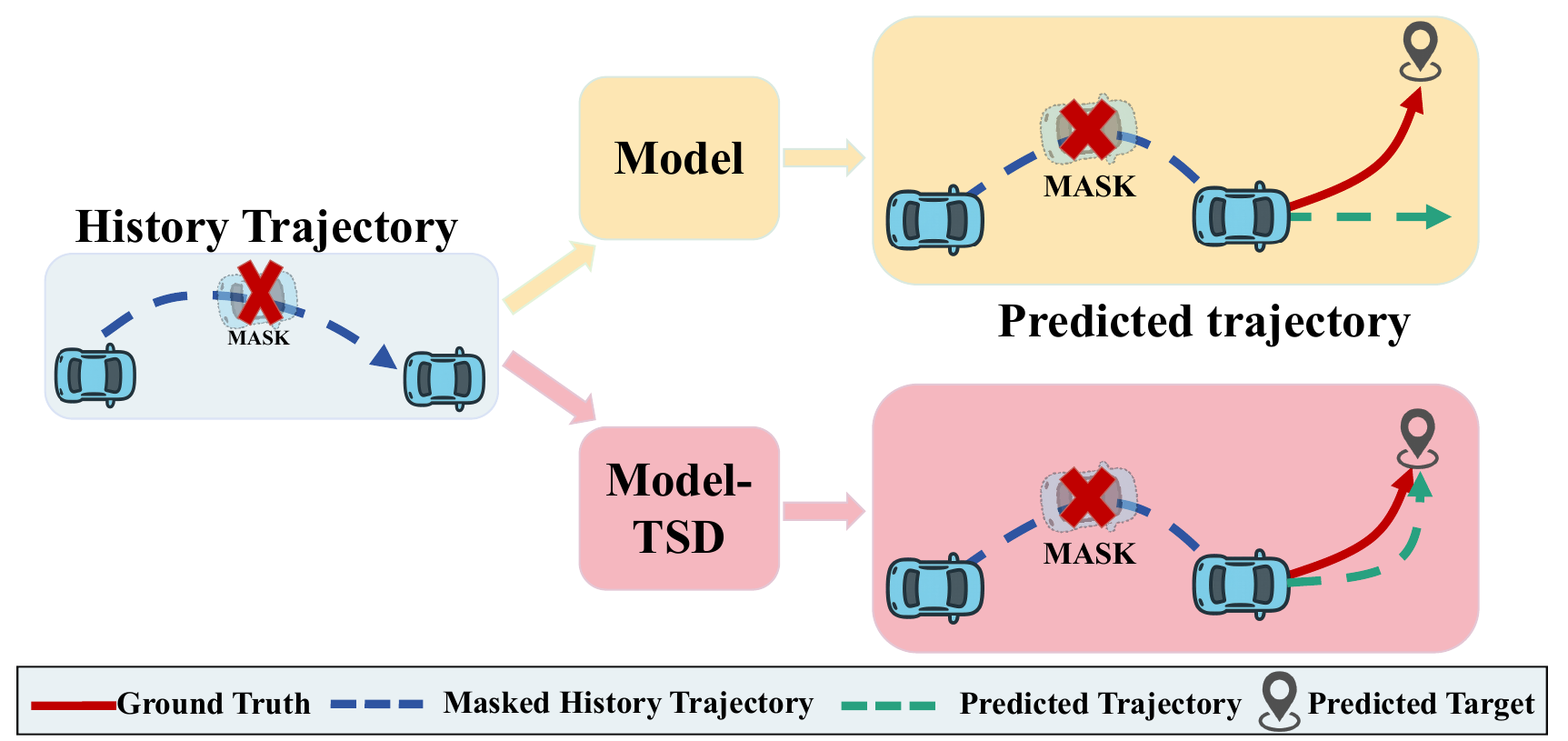}

\caption{Motion forecasting models often encounter occluded historical trajectories. Our proposed TSD enhances robustness by leveraging self-distillation and predicting target to adapt to Partially Observation  scenarios.}
\label{fig:motivation}

\end{figure}

Many carefully designed networks have achieved excellent results in motion forecasting, including recently proposed methods such as HiVT~\cite{hivt},DeMo~\cite{zhang2024decoupling}, and LAformer~\cite{liu2024laformer}, which have performed exceptionally well on large-scale motion forecasting datasets like Argoverse 1,Argoverse2 and Nuscenes. Despite their impressive performance, most open-source large-scale motion forecasting datasets contain carefully collected agent motion data, where the agents typically have complete input and output observations. However, in real-world deployments, the agent observation inputs received by trajectory prediction models typically originate from upstream objection tracking models. Due to interference from road obstacles and occlusions by dynamic vehicles, tracking results are highly prone to objection loss and localization errors, resulting in incomplete agent observation inputs. These inputs often exhibit random frame drops or partial trajectory absences, leading to erroneous trajectory predictions and consequently compromising the safety of autonomous driving systems.

Some methods~\cite{xu2024adapting,sun2022human,qi2025robust,qi2025action} have designed specific networks to address the challenges of partial observation. However, if the goal is to retain the knowledge of the state-of-the-art network trained on fully observed data, many approaches adopt knowledge distillation~\cite{monti2022many,pop,wang2024adversarial,ye2025safedriverag}. Specifically, a network is first trained on fully observed data, and then the knowledge from this network is distilled into an identical network that receives only partially observed inputs. While the network obtained through knowledge distillation retains the original model's structure and performs well under partial observation, this approach has two main drawbacks: first, the performance of the distilled network on fully observed data is lower than that of the originally trained network; second, the knowledge distillation process is cumbersome, requiring the training of one network followed by another distillation step, making the entire training process very expensive.

To overcome the aforementioned challenges, as depicted in Figure~\ref{fig:motivation} we propose a novel \textbf{T}arget-driven \textbf{S}elf-\textbf{D}istillation method (\textbf{TSD}) for motion forecasting. First, when observations are incomplete, if accurate target points can guide the trajectory prediction process, the model can still make accurate predictions despite the limited observations. Therefore, we propose an anchor-free target point generation method that accurately predicts the target points of agents, thereby guiding accurate trajectory prediction under both partial and full observations. Second, since partially observed trajectories can be viewed as a natural data-distortion branch, we refer to the design for classification tasks in~\cite{xu2019data,deng2025global,lv2025t2sg,zhu2023unsupervised} and develop a self-distillation method for motion forecasting. This method uses fully observed and partially observed inputs as two branches, jointly optimizes these branches, and employs empirical Maximum Mean Discrepancy (MMD)~\cite{mmd} as a non-parametric metric to measure the consistency of feature distributions between these distorted versions. Through these two aspects, we can obtain a robust model for partially observed data, and through an end-to-end training process, its performance is comparable to or even surpasses the original model under full observation conditions.

Our contributions can summarized as follows:

\par\textbf{(1)}~We propose \textbf{TSD}, a novel plug-and-play method which enhances the robustness of motion prediction models under partially observed trajectories while maintaining or even improving their motion prediction capabilities under fully observed trajectories.

\par\textbf{(2)}~We design an anchor-free target point generation method based on the Transformer decoder. By accurately predicting target points, this method guides the model in performing precise trajectory prediction under partial observation conditions.

\par\textbf{(3)}~We introduce a new self-distillation mechanism for partially observed trajectories task. This enhances the robustness of the model under partial observation conditions.

\par\textbf{(4)}~Experimental results on multiple large-scale datasets demonstrate that our approach not only significantly improves baseline robustness under partially observed scenarios, but also enhances the performance of some baselines in fully observed conditions, achieving comprehensive performance gains.

\section{Related Works}

\textbf{Fully Observed Motion Forecasting} Fully Observed Motion Forecasting typically uses complete historical trajectories to predict future motion. Many methods have been proposed in current research, which can be categorized into anchor-based and anchor-free approaches. Methods like \cite{chai2019multipath},\cite{zhao2021tnt}, and \cite{afshar2024pbp} cluster representative trajectory anchors as static anchors using unsupervised learning from training data. On the other hand, \cite{varadarajan2022multipath++}, \cite{nayakanti2023wayformer}, \cite{shi2022motion}, and \cite{ngiam2022scenetransformerunifiedarchitecture} generate dynamic, learnable anchors through implicit methods. After anchor generation, these methods combine the anchors with contextual information from the scene and use simple networks to predict future trajectories.  Anchor-free methods, such as those using diffusion models like \cite{jiang2023motiondiffuser}, \cite{li2023bcdiff}, and \cite{gu2022stochastic}, or generative networks like \cite{barsoum2018hp}, \cite{kingma2013auto}, and \cite{gomez2020real}, rely on the precise capturing of multimodal distributions for prediction. Additionally, some anchor-free methods adopt self-supervised learning approaches, such as \cite{chen2023traj}, \cite{cheng2023forecast} and \cite{lan2023sept}. These methods learn more comprehensive scene interaction information by reconstructing features from masked regions and can also achieve impressive trajectory forecasting results. However, these methods, trained on carefully designed datasets with complete trajectory observations, exhibit poor robustness to partially observed trajectory inputs. Their performance typically degrades significantly when encountering those cases. Our proposed plug-and-play method TSD can significantly enhance the robustness of these approaches when facing partially observed trajectories and maintain or even improve their performance under fully observed inputs.

\begin{figure*}[t]
    \centering
    \includegraphics[width=0.9\textwidth]{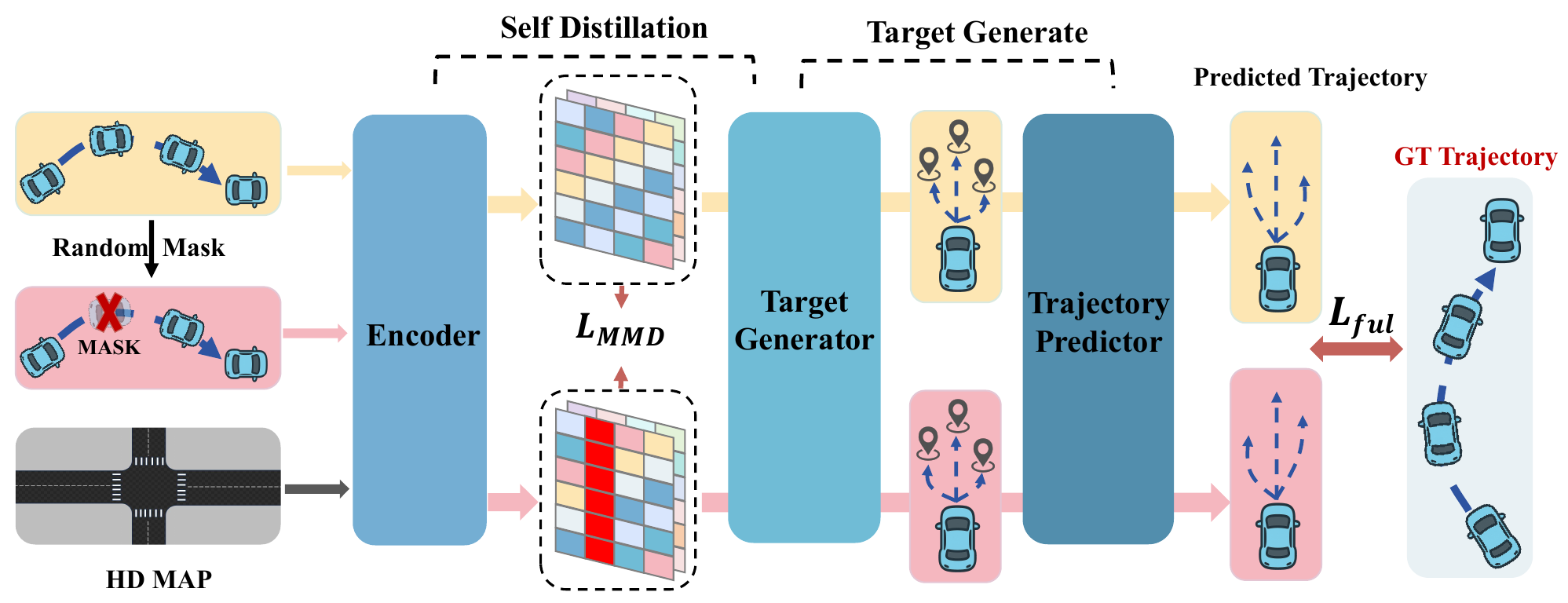}
    
    \caption{Overview of our proposed TSD. The entire training process is implemented end-to-end. We first apply random masking to the input trajectories to obtain partially observed trajectory branches, which are then fed into the network along with fully observed trajectories. Our proposed target generator produces sequential targets, which subsequently guide the trajectory prediction process. The features extracted from partially observed trajectories and fully observed trajectories are brought closer in distribution using MMD loss. The generation of partially observed trajectories occurs only during training, based on the original fully observed trajectories.}
    \label{fig:overview}
    
\end{figure*}

\textbf{Partially Observed Motion Forecasting}~Partially Observed Motion Forecasting predicts future trajectories from Partially historical data. Fully Observed Motion Forecasting methods perform well on complete trajectories but their accuracy drops sharply when parts of the historical data are masked out.Numerous works have been proposed to address this issue. Some methods attempt to utilize knowledge distillation to mitigate this problem. \cite{monti2022many} and \cite{pop} use a teacher model trained with full trajectories to guide the student model, enhancing its prediction capability under Partially observation conditions. \cite{li2024lakd} introduces a knowledge distillation framework for dynamically varying trajectory lengths and a dynamic neuron soft-masking mechanism to improve the model’s predictive performance when dealing with historical trajectory data of different lengths. Additionally, \cite{xu2024adapting,qi2020stc} learns representations from trajectory data of varying observation lengths and generates time-invariant representations, which are further optimized through an adaptive mechanism, thus improving prediction accuracy. \cite{li2023bcdiff} uses a bidirectional consistency diffusion model to predict future pedestrian trajectories based on short historical trajectory segments. \cite{sun2022human,qi2021semantics} proposes a unified feature extractor and a novel pre-training mechanism to capture relevant information in momentary observations, achieving pedestrian trajectory prediction even with extremely short observation lengths. However, existing methods introduce additional training costs and only address scenarios with continuous missing input trajectories. In contrast, our proposed TSD method achieves end-to-end model training through self-distillation, significantly reducing computational overhead while effectively handling both continuous and random missing trajectory scenarios, demonstrating stronger robustness.

\section{Problem Definition}

In the task of motion forecasting, the objective is to predict the future state sequence 
$X_{\text{pred}} = \{s_t\}_{t=1}^{T_{\text{pred}}}$ of a given agent, based on its past state sequence $X_{\text{obs}} = \{s_t\}_{t=-T_{\text{obs}}+1}^0$, where $s_t$ represents the state of the agent to be predicted at time $t$. During the prediction, we denote the states of surrounding agents as $A$, and elements and polylines from the high-definition map provided by the dataset, including lane centerlines, drivable areas, and so on, are denoted as $M$. Note that the quantities of $A$ and $M$ may not be fixed in a given context. Therefore, our overall objective is to obtain a conditional probability distribution $p(X_{\text{pred}}|X_{\text{obs}}, A, M)$.

Due to the diversity of agents' motion, the conditional probability distribution $p(X_{\text{pred}}|X_{\text{obs}}, A, M)$ can be highly multimodal (\textit{e.g.}, left turn, right turn, going straight, stopping, etc.). By decomposing this task into a target point prediction issue, in the case of predicting trajectories for $K$ modes, the probability distribution can be decomposed as follows:
\begin{align}
    p(X_{\text{pred}}|X_{\text{obs}}, A, M) &= \sum_{k=0}^{K} p(\tau_k|X_{\text{obs}}, A, M) \nonumber \\
    &\quad \cdot p(X_{\text{pred}}|X_{\text{obs}}, A, M, \tau_k),
\end{align}
where $\tau_k$ represents the predicted target point corresponding to the $k^{th}$ mode.

Note that when it comes to partial motion forecasting, the initial observation of past state sequence ( $X_{\text{obs}}$ ) and the agent states ( A ) are not available at every frame;After our masking process, only a portion of the observed data is input into the motion prediction model. We categorize the partial trajectory masking of the target agent in the motion prediction task into two masking patterns:
(1) Random Masking: This involves randomly selecting and masking certain frames in the complete historical sequence to simulate discrete trajectory gaps caused by objection loss during the operation of a objection tracking model.
(2) Continuous Masking: This involves continuously masking the complete historical trajectory starting from the first frame to simulate scenarios where the target agent is occluded for an extended period, resulting in the absence of an entire trajectory segment.
\section{Method}
\subsection{Overview}
The overview of our proposed framework is illustrated in Figure~\ref{fig:overview}. We first use an encoder to extract representations of the target agent, surrounding agents, polylines, and elements on the high-definition map. Subsequently, we use a sequential target point generator to iteratively predict long-term and short-term target points for the interested agent over different time horizons. These target points are then used to guide the trajectory predictor to iteratively predict multi-modal trajectories. Additionally, during the training phase, we add a partial observation branch. By generating masks, we create partially observed trajectories and feed them into the model along with fully observed trajectories for inference. The inference results from the partially observed trajectories are backpropagated in the same manner as those from fully observed trajectories, and the features extracted from partially observed trajectories are constrained by the features from fully observed trajectories through a loss function.

\subsection{Encoder}
\noindent For motion forecasting, the first step is to use an encoder to embed the trajectory features of each agent and the features of the high-definition map elements, while also exploiting the interactions among agents and map elements. Given $X_{\text{obs}}, A, M$, the encoder outputs the feature embedding $F_X = \{F_{s_t}\}_{t=-T_{\text{obs}}+1}^0 \in \mathbb{R}^{T_{\text{obs}} \times d}$ of the interested agent at historical steps, where $d$ represents the size of the hidden dimension. Additionally, the encoder outputs the features $F_A$ of surrounding agents, the features $F_M$ of map elements, the interaction features $F_{X \rightarrow A}$ between agents and the interaction features $F_{X \rightarrow M}$ between agents and map elements.

\subsection{Target Generation and Trajectory Prediction}
\noindent Here, we propose a target point generation method based on the Transformer~\cite{vaswani2017attention} decoder, which can iteratively generate sequential target points over both long-term and short-term time horizons. 

Inspired by~\cite{zhou2023query}, we assume that the $x$ and $y$ coordinates of target points $\tau_k^i$ of the $k^{th}$ mode to be predicted, where $i \in \{1, 2, \cdots, n\}$, follow a Laplace distribution, denoted as $\tau_{k,x}^i, \tau_{k,y}^i \sim \mathcal{L}(\mu, b)$. Here, $n$ is the number of target points we set artificially for long-time horizon generation. Therefore, in our target point generation, the ultimate goal is to predict the parameters $\mu_{k,x}^i,\mu_{k,y}^i$ and $b_{k,x}^i,b_{k,y}^i$ for the $x$ and $y$ coordinates of each target point of every mode. Given the multimodal prediction of trajectories for $K$ modes, we generate $n$ target points for each mode; thus, the final set of target points generated for each agent is $\tau \in \mathbb{R}^{K \times n \times 4}$. For the dimension $n$, target points are generated sequentially over $n$ cycles, each time using the information gathered so far, rather than through an auto-regressive process. Eventually, we can concatenate them as a sequence. 

To achieve this, we employed three cross-attention blocks, allowing the target queries to interact and exchange information with the map features, surrounding agents' features, and the historical features of the interested agent. During the information exchange with each type of feature, we also separately utilized the interaction features between the interested agent and the map, surrounding agents, denoted as $F_{X \rightarrow M}$ and $ F_{X \rightarrow A} $, as well as the interaction features between different time points of the interested agent, denoted as $F_{s_i \rightarrow s_j}$, where $i, j \in \{T_{\text{obs}+1}, \cdots, -1, 0\}$.

Subsequently, we employ two three-layer Feed-Forward Neural Networks (FFNs) to separately deduce the parameters $\mu_{k,x}, \mu_{k,y}$ and $b_{k,x}, b_{k,y}$ of the Laplace distribution for the x and y coordinates of each target points' mode. Based on the predetermined number of target points $N$, in each iteration, the predicted target point is positioned relative to the previously predicted target point or the starting point, at a future time step of $T_{\text{pred}}/N$. To optimize our target point predictions, we utilize the negative log-likelihood of the Laplace distribution:
\begin{align}
    \mathcal{L}_{\text{tar}} &= \sum_{i=1}^{N} \sum_{j \in \{x,y\}} \sum_{k=1}^{K} -\log\left(\mathcal{L}(s_{\tau_i};\mu_{k,j}^i,b_{k,j}^i)\right) \nonumber \\
    &= \sum_{i=1}^{N} \sum_{j \in \{x,y\}} \sum_{k=1}^{K} \left(\log(2b_{k,j}^i) + \frac{|s_{\tau_i}-\mu_{k,j}^i|}{b_{k,j}^i}\right),
    \label{eqo:target}
\end{align}
where $s_{\tau_i}$ represents the actual state of the agent at the time step corresponding to the $i^{th}$ generated target point, and
\begin{equation}
    \mathcal{L}(s_{\tau_i};\mu_{k,j}^i,b_{k,j}^i)=\frac{1}{2b_{k,j}^i} \exp\left(-\frac{|s_{{\tau_i}}-\mu_{k,j}^i|}{b_{k,j}^i}\right).
\end{equation}

Once obtaining embeddings of the target point $e_{\tau}$, we use them to guide the prediction of future trajectories. For each target point embedding $e_{\tau_i}$, where $i \in \{1, 2, \cdots, n\}$, it will guide the trajectory prediction from its previous target point or starting point to the time step corresponding to that target point embedding, with a length of $T_{\text{pred}} / N$.

Similar to target point generation, we set $K$ trajectory queries for each agent, which correspond to the target embeddings derived from each target point query. The trajectory queries first interact with map features, agent features, and historical trajectory features using the cross-attention module, and then we feed them into a multi-head attention module. The query, key, and value are computed as follows:
\begin{align}
    Q=W_q (e_T + e_{\tau_i}), \quad K=W_k (e_T + e_{\tau_i}), \quad V=W_v e_T, \nonumber
\end{align}
where $e_T$ is the embedding of the trajectory query. Finally, we perform self-attention interactions among different modes of the same agent. We also treat the $x$ and $y$ coordinates of each step of an agent's trajectory as following a Laplace distribution like the target point generation, and decode the location and scale of the future trajectory using two FFNs. This process will be repeated $N$ times to generate the complete predicted trajectory. Then, we use a mixture factor $\pi$ to assign probabilities to each predicted trajectory mode.

We employ two loss functions to optimize the trajectory prediction and the mixture coefficients. For the optimization of trajectory prediction, the loss function is similar to that of Equation~\ref{eqo:target}, with the difference being that it calculates the result of each step in trajectory prediction. We denote this loss as $\mathcal{L}_{\text{reg}}$. During the optimization of the mixture coefficients, the position and scale of the trajectory are detached from the computational graph, focusing solely on optimizing the mixture coefficients:
\begin{equation}
    \mathcal{L}_{\text{cls}}=-\log(\sum_{k=1}^{K} \pi_k \sum_{i=1}^{T_{\text{pred}}}(-\log(2b_{k,i}) + \frac{|s_{k,i}-\mu_{k,i}|}{b_{k,i}})),
    \label{eqo:cls}
\end{equation}
where $K$ represents the total number of modes to be predicted, and $\pi_k$ is the probability score assigned to the $k^{th}$ mode.

\subsection{Self-Distillation}

To end-to-end train a model that is robust to partially observed trajectory inputs, we have designed a self-distillation method specifically for the motion prediction task. The detailed process of this method is as follows:\\
\textbf{(1)}~The input trajectory data are divided into two branches: complete observations and partial observations. The partially observed trajectory data are obtained by applying random masking to the data $X_{obs}$ and $A$ from the complete observation branch.\\
\textbf{(2)}~Both the complete and partially observed data are fed into the motion prediction network to obtain the intermediate output features and the final prediction results.\\
\textbf{(3)}~For the representation vectors of the intermediate features, we use Maximum Mean Discrepancy (MMD)~\cite{mmd} to reduce the differences between them. To match the feature distributions, a metric between the representation vectors of the two-branch distorted versions needs to be defined. We adopt the empirical MMD~\cite{mmd} as a non-parametric metric that has been widely used in domain adaptation to measure the discrepancy of distributions. We use the following empirical MMD loss to minimize the margin between the feature distribution obtained by full observation trajectories ($p(F_A), p(F_X)$ and $p(F_{X\rightarrow A})$) with feature distribution obtained by partial observation trajectories ($p(F_A^P), p(F_X^P)$ and $p(F_{X\rightarrow A}^p)$) :
\begin{equation}
\begin{aligned}
    \mathcal{L}_{\text{MMD}}= &|| \frac{1}{n} \sum_{i=1}^n \text{Concat}(F_{Xi}, F_{Ai}, F_{X\rightarrow Ai}) \\ 
    &- \frac{1}{n} \sum_{i=1}^n \text{Concat}(F_{Xi}^P, F_{Ai}^P, F_{X\rightarrow Ai}^P)||_2^2,
\end{aligned}
\label{eqo:mmd}
\end{equation}
where $n$ denotes the number of total features, $F_A, F_A^P$ denote features of surrounded agents, $F_X, F_X^P$ are features of the interested agent, and $F_{X\rightarrow A}, F_{X\rightarrow A}^P$ are the interaction features between the interested agent and surrounded agents.
MMD is adopted instead of KL divergence or L2 distance due to its theoretical and practical advantages in our self-distillation framework. As a kernel-based, non-parametric measure, MMD captures distributional discrepancies without assuming any parametric form, making it suitable for our task where the feature distributions of fully and partially observed trajectories are often non-Gaussian, multi-modal, and partially disjoint. In contrast, KL divergence is asymmetric and unstable when the distributions have non-overlapping support, while L2 distance overlooks the geometric structure of the feature space, leading to inferior alignment. Prior studies~\cite{huang2017like,xu2019data} have also verified MMD’s stability and effectiveness in knowledge distillation, confirming it as a theoretically grounded and empirically robust choice for aligning latent representations.

\textbf{(4)}~For both the complete and partially observed prediction results, the aforementioned $\mathcal{L}_{tar}, \mathcal{L}_{reg}$ and $\mathcal{L}_{cls}$ are used for optimization. 

Compared to the training of standard baseline networks, self-distillation training does not add any additional network parameters.

\subsection{Training and Inference}

We propose that the self-distillation process is only used during training. Specifically, during the training process, we adopt a winner-takes-all training strategy~\cite{lee2016stochastic,qi2019attentive}. We select the trajectory with the best final target point prediction result from the complete observation for back-propagation and choose the trajectory with the same index from the partial observation output results. Overall, we denote loss for the full observation input branch as $\mathcal{L}_{ful}=\mathcal{L}_{tar}+\mathcal{L}_{reg} + \mathcal{L}_{cls}$, and the loss for the partial trajectory input branch is denoted as $\mathcal{L}_{par}=\mathcal{L}_{tar}+\mathcal{L}_{reg} + \mathcal{L}_{cls}$. Our overall training loss is as follows:
\begin{equation}
    \mathcal{L} = \mathcal{L}_{ful} + \mathcal{L}_{par} + \mathcal{L}_{MMD} 
\end{equation}
\begin{table*}[!t]
  \centering
  \resizebox{\textwidth}{!}{
  \begin{tabular}{c|c|ccccc} 
    \toprule
    Dataset & Method & Fully Obs.  & Mask Rate. = 20\% & Mask Rate. = 40\% & Mask Rate. = 60\%  & Mask Rate. = 80\% \\ \midrule
    \multirow{3}{*}{Argoverse 1} & HiVT-64 & 0.687/1.030/0.102 & 
    0.735/1.113/0.120 & 
    0.793/1.207/0.138 & 
    0.859/1.331/0.167 & 
    0.950/1.489/0.200  \\
    & HiVT-POP-64 & 
    0.693/1.043/0.103 & 
    0.732/1.103/0.121 & 
    0.790/1.199/0.137 & 
    0.842/1.283/0.155 & 
    0.894/1.361/0.171  \\
    & HiVT-TSD-64 & 
    \textbf{0.675}/\textbf{0.998}/\textbf{0.096}
    &
    \textbf{0.698}/\textbf{1.038}/\textbf{0.103} & \textbf{0.722}/\textbf{1.077}/\textbf{0.112} & \textbf{0.761}/\textbf{1.141}/\textbf{0.123} & \textbf{0.837}/\textbf{1.256}/\textbf{0.146}  \\ \midrule
    \multirow{3}{*}{Argoverse 1} & HiVT-128 & 0.661/0.969/0.092 & 
    0.713/1.059/0.107 & 
    0.783/1.188/0.135 & 
   0.860/1.330/0.162 & 
   0.931/1.454/0.189  \\
    &  HiVT-POP-128 & 
    0.672/0.989/0.096 & 
    0.713/1.066/0.109 & 
    0.760/1.144/0.123 & 
    0.812/1.233/0.144 & 
    0.873/1.334/0.165 
    \\
    & HiVT-TSD-128 & \textbf{0.653}/\textbf{0.944}/\textbf{0.088} & \textbf{0.700}/\textbf{1.029}/\textbf{0.102} & \textbf{0.716}/\textbf{1.057}/\textbf{0.113} & \textbf{0.758}/\textbf{1.143}/\textbf{0.127} & \textbf{0.832}/\textbf{1.235}/\textbf{0.151} \\
    \midrule

  \end{tabular}
  }
\resizebox{\textwidth}{!}{
  \begin{tabular}{c|c|ccccc} 
    \toprule
   Dataset & Method & Fully Obs.  & Mask Rate. = 20\% & Mask Rate. = 40\% & Mask Rate. = 60\%  & Mask Rate. = 80\% \\ \midrule
    \multirow{4}{*}{Argoverse 2} & DeMo & 
    0.680/1.354/0.155 & 
    1.199/2.132/0.268 & 
    1.976/3.352/0.372 & 
    1.999/3.146/0.379 & 
    2.197/3.429/0.416  \\
    & DeMo-pop & 
    0.685/1.349/0.156 & 
    0.913/1.648/\textbf{0.218} & 
    1.527/2.694/0.359 & 
    2.447/4.217/0.460 & 
    2.399/4.032/0.449  \\
    & Forecast-mae & 
    0.712/1.408/0.217 & 
    1.488/2.729/0.354 & 
    2.703/4.957/0.511 & 
    3.719/6.761/0.579 & 
    3.655/6.529/0.562  \\
    & DeMo-TSD &     \textbf{0.673}/\textbf{1.303}/\textbf{0.155} & \textbf{0.821}/\textbf{1.527}/0.460
    &\textbf{1.143}/\textbf{2.119}/\textbf{0.308} & \textbf{1.423}/\textbf{2.584}/\textbf{0.368} & \textbf{1.172}/\textbf{2.027}/\textbf{0.311}
    \\ \midrule

  \end{tabular}

  }

\resizebox{\textwidth}{!}{
  \begin{tabular}{c|c|ccccc} 
    \toprule
    Dataset & Method & Fully Obs.  & Mask Rate. = 20\% & Mask Rate. = 40\% & Mask Rate. = 60\%  & Mask Rate. = 80\% \\ \midrule
    \multirow{3}{*}{Nuscenes} & LAformer & \textbf{1.197}/\textbf{2.292}/\textbf{0.365} & 
    1.356/2.422/0.379 & 
    1.565/2.611/0.405 & 
    1.740/2.793/0.424 & 
    2.001/3.163/0.456  \\
    & LAformer-pop & 
    1.235/2.350/0.376 & 
    1.248/\textbf{2.376}/0.382 & 
    1.267/2.405/0.386 & 
    1.298/2.542/0.391 & 
    1.421/2.589/0.409 \\
    & LAformer-TSD & 1.220/2.361/0.367 & \textbf{1.237}/2.378/\textbf{0.375} & \textbf{1.256}/\textbf{2.393}/\textbf{0.379} & \textbf{1.294}/\textbf{2.460}/\textbf{0.386} & \textbf{1.340}/\textbf{2.534}/\textbf{0.397}
    \\ \midrule

  \end{tabular}

  }

  \caption{
  We conducted comprehensive comparative experiments on three mainstream motion prediction benchmark datasets (Argoverse 1, Argoverse 2, and NuScenes) to evaluate the prediction performance of the original model and its enhanced versions using the POP and TSD methods under partially observed conditions with random masking. The experiments employed three core metrics—\textbf{minADE$_k$}/\textbf{minFDE$_k$}/\textbf{MR$_k$} for evaluation, with k set to 6 for the Argoverse 1 and Argoverse 2 datasets, and k set to 5 for the NuScenes dataset. The best-performing results in all comparisons are highlighted in \textbf{bold}.
}
  \label{tab:robustness}
  
\end{table*}

\begin{table*}[!htbp]
  \centering
  \small
  \resizebox{\textwidth}{!}{
  \begin{tabular}{c|c|ccccc} 
    \toprule
    Dataset. & Method & Fully Obs. & Obs. = 15 & Obs. = 10 & Obs. = 5 & Obs. = 1 \\ \midrule
    \multirow{3}{*}{Argoverse 1} & HiVT-64 & 
    0.687/1.030/0.102 & 
    0.774/1.168/0.132 &
    0.720/1.085/0.114 & 
    0.703/1.061/0.108 & 
    1.120/1.761/0.253  
    \\
    & HiVT-POP-64 & 
    0.693/1.043/0.103 &
    0.692/1.040/0.104 & 
    0.704/1.060/0.107 &
    0.742/1.117/0.118 &
    1.228/1.857/0.227 
    \\
    & HiVT-TSD-64 & 
    \textbf{0.675}/\textbf{0.998}/\textbf{0.096} & 
    \textbf{0.679}/\textbf{1.009}/\textbf{0.098}
    & 
    \textbf{0.694}/\textbf{1.031}/\textbf{0.104} & 
    \textbf{0.739}/\textbf{1.098}/\textbf{0.117} & 
    \textbf{1.040}/\textbf{1.600}/\textbf{0.211}
    \\ \midrule
    \multirow{3}{*}{Argoverse 1} & HiVT-128 & 
    0.661/0.969/0.092 &
    0.681/1.005/0.098 &
    0.693/1.023/0.101 &
    0.742/1.098/0.116 &
    1.056/1.652/0.231
    \\
    & HiVT-POP-128 & 
    0.672/0.990/0.096 &
    0.675/1.017/0.101 &
    0.688/1.040/0.106 &
    0.742/1.105/0.119 &
    1.062/1.675/0.237
    \\
    & HiVT-TSD-128 & 
    \textbf{0.654}/\textbf{0.957}/\textbf{0.090}
    & 
    \textbf{0.668}/\textbf{0.983}/\textbf{0.093}
     & \textbf{0.683}/\textbf{1.009}/\textbf{0.100} & 
    \textbf{0.730}/\textbf{1.076}/\textbf{0.111}
    & 
    \textbf{1.002}/\textbf{1.516}/\textbf{0.191}
 \\
    \bottomrule
  \end{tabular}}

  \caption{Comparison the \textbf{minADE$_6$}/\textbf{minFDE$_6$}/\textbf{MR$_6$} performance of HiVT, HiVT-POP, and HiVT-TSD under Continuous Masking partial observation trajectory inputs, conducted on the Argoverse motion forecasting dataset. the best results are highlighted in \textbf{bold}.}
  \label{tab:table2}
  \end{table*}
\section{Experiments}

\begin{figure*}[!htbp]
\includegraphics[width=0.88\textwidth]{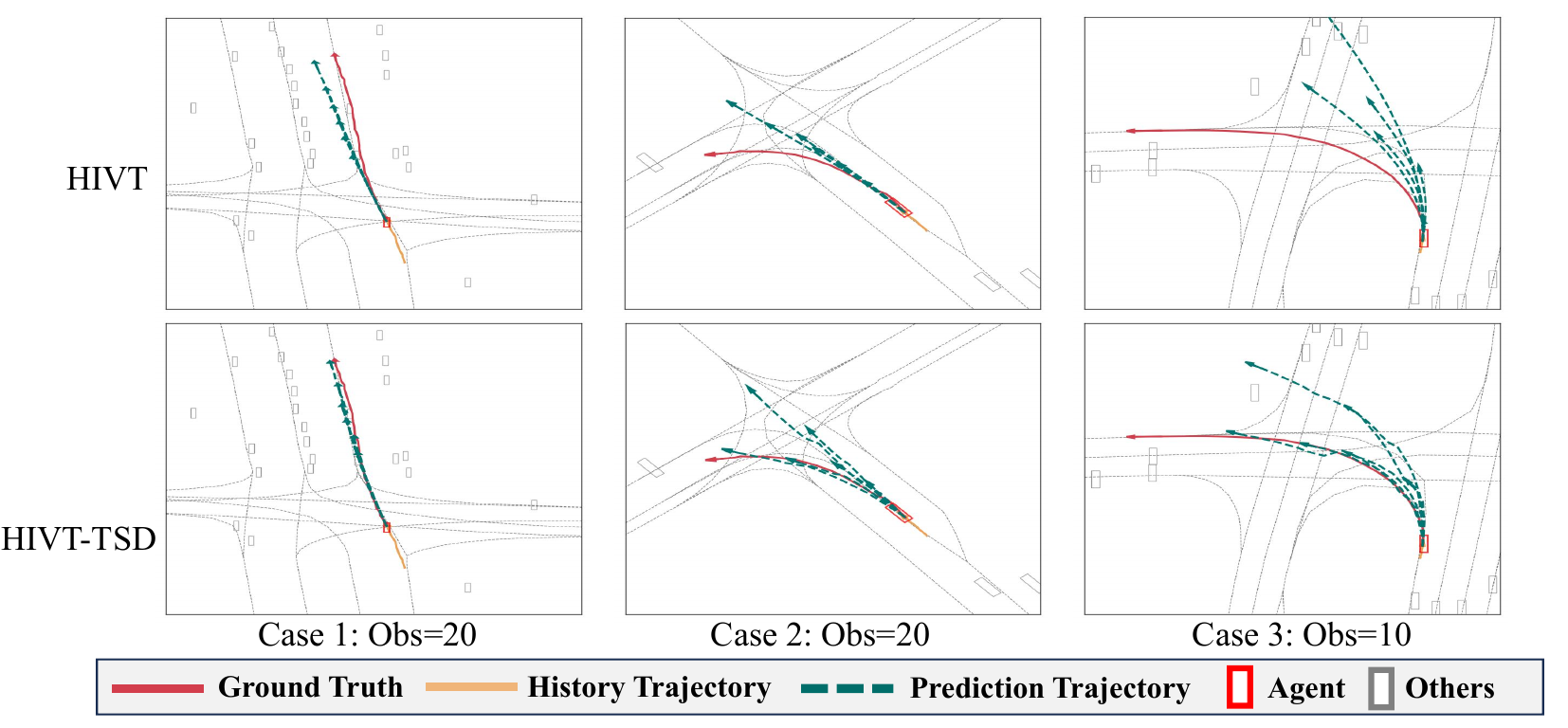}
\centering
\caption{Qualitative results of HiVT and HiVT-TSD. The past trajectories are shown in yellow, the ground-truth trajectories are shown in red, and the predicted trajectories are shown in green. The white boxes are other vehicles around, and the red boxes are agents}
\label{fig:subfigures}
\end{figure*}

\subsection{Experiment Settings}

\noindent\textbf{Dataset.}~We evaluated our model on the Argoverse1~\cite{argoverse1}, Argoverse2~\cite{wilson2023argoverse}, and nuScenes~\cite{caesar2020nuscenes} datasets. Argoverse1 contains 323,557 driving scenarios (205,942/39,472/78,143 for train/val/test) with 5s sequences sampled at 10Hz. Models observe the first 2s and predict the next 3s. Argoverse2 provides 249,880 scenarios (199,908/24,988/24,984 for train/val/test) sampled at 10Hz, requiring 6s prediction based on the previous 5s. nuScenes includes 49,787 8s scenes (32,186/8,560/9,041 for train/val/test) at 2Hz, with 2s history to predict 6s future trajectories. All datasets provide agent trajectories and high-definition maps.

\noindent\textbf{Metrics.} We adopted evaluation metrics commonly used in previously state-of-the-art methods~\cite{zhou2023query,wang2023prophnet,feng2023macformer}, \textit{i.e.}, minimum Average Displacement Error (\textbf{minADE}$_K$), minimum Final Displacement Error (\textbf{minFDE}$_K$), and Miss Rate (\textbf{MR}$_K$). The minADE evaluates the $L2$ distance between the coordinates of each step of the top K best-predicted future trajectories by the model and the ground truth, averaged over all time steps. The minFDE evaluates the accuracy of the last point's predicted coordinates. The MR metric evaluates the percentage of predictions in minFDE that exceed 2 meters.Only the top K predicted trajectories are evaluated if the model outputs more than K trajectories.For the Argoverse 1 and Argoverse 2 datasets,  K  is set to 6, while for the nuScenes dataset,  K  is set to 5.

\noindent\textbf{Implement Details.}~We trained HiVT, DeMo, LAformer and Forecast-MAE with our method using PyTorch on 8 RTX 4090 GPUs, following the original hyperparameter settings. 
\begin{table*}[t]
  \centering
  \small
  \setlength\tabcolsep{2mm} 
  \begin{tabular}{cc|ccccc}
    \toprule
    Target & SD & Obs. = 1 & Obs. = 5 & Obs. = 10 & Obs. = 15 & Obs. = 20 \\
    \midrule
    & & 1.120/1.761/0.253 & 0.774/1.168/0.132 & 0.720/1.085/0.114 & 0.703/1.061/0.108 & 0.687/1.030/0.102 \\
    $\checkmark$ & & 1.089/1.706/0.248 & 0.754/1.130/0.124 & 0.711/1.065/0.109 & 0.696/1.041/0.103 & 0.677/1.002/0.099 \\
    $\checkmark$ & $\checkmark$ & \textbf{1.040}/\textbf{1.600}/\textbf{0.211} & \textbf{0.739}/\textbf{1.098}/\textbf{0.117} & \textbf{0.694}/\textbf{1.031}/\textbf{0.104} & \textbf{0.679}/\textbf{1.009}/\textbf{0.098} & \textbf{0.675}/\textbf{0.998}/\textbf{0.096} \\
    \bottomrule
  \end{tabular}
  \caption{Comparison of the \textbf{minADE$_6$}/\textbf{minFDE$_6$}/\textbf{MR$_6$} performance of different ablation versions of HiVT-TSD. The best results are highlighted in \textbf{bold}.}
  \label{tab:comp_ab}
\end{table*}

\begin{table*}[!htbp]
  \centering
  \small
  \setlength\tabcolsep{2mm} 
  \begin{tabular}{cc|ccccc}
    \toprule
    two-branch & MMD & Obs. = 1 & Obs. = 5 & Obs. = 10 & Obs. = 15 & Obs. = 20 \\
    \midrule
    & & 1.050/1.625/0.218 & 0.754/1.130/0.124 & 0.711/1.065/0.109 & 0.696/1.041/0.103 & 0.677/1.002/0.099 \\
    $\checkmark$ & & 1.089/1.706/0.248 & 0.740/1.106/0.118 & 0.696/1.040/0.105 & 0.681/1.014/0.100 & 0.677/1.005/0.098 \\
    $\checkmark$ & $\checkmark$ & \textbf{1.040}/\textbf{1.600}/\textbf{0.211} & \textbf{0.739}/\textbf{1.098}/\textbf{0.117} & \textbf{0.694}/\textbf{1.031}/\textbf{0.104} & \textbf{0.679}/\textbf{1.009}/\textbf{0.098} & \textbf{0.675}/\textbf{0.998}/\textbf{0.096} \\
    \bottomrule
  \end{tabular}
  \caption{Comparison of the \textbf{minADE$_6$}/\textbf{minFDE$_6$}/\textbf{MR$_6$} performance of different ablation versions of self-distillation part of HiVT-TSD. The best results are highlighted in \textbf{bold}.}
  \label{tab:sd_ab}
\end{table*}
\subsection{Result and Quantitative Comparisons}
\noindent\textbf{Robustness with Random Masking: } We systematically evaluated the motion prediction performance of the proposed method across multiple datasets. As shown in Table~\ref{tab:robustness}, we developed three enhanced versions—HiVT-TSD, DeMo-TSD, and LAformer-TSD—based on the mainstream model frameworks HiVT, DeMo, and LAformer, respectively, and conducted comparative experiments with the state-of-the-art knowledge distillation-based method, POP.In addition, we include Forecast-mae in our comparison as a representative mask-based self-supervised pretraining method.

The experimental results demonstrate that as the mask ratio increases under partially observed data with random masking, the prediction performance of all models significantly declines. Notably, while the POP method yields some performance improvements over the original models under random masking conditions, it comes at the cost of reduced prediction accuracy under full observation. Moreover, in certain model architectures, the performance under random masking is even worse than that of the original models (LAformer-POP at masking rates of 60\%–80\%). In contrast, our proposed TSD method not only significantly outperforms the POP method under high mask ratios but also fully preserves or even slightly enhances the original model performance under complete observation data.On the Argoverse 2 dataset, the results of Forecast-mae show that its performance drops sharply as the masking ratio increases. This indicates that although Forecast-mae benefits from learning structured masking-trajectory reconstruction during pretraining, the learned priors fail to generalize well to the random masking patterns encountered at inference time, resulting in limited robustness under severe observation loss.

To further verify the reliability of these improvements, we conducted paired \emph{t}-tests across five masking ratios for each model. The results show that the improvements of HiVT (p = 0.0241) and DeMo (p = 0.0336) are statistically significant, while LAformer exhibits a similar trend (p = 0.0656 $\approx$ 0.05). These analyses confirm that the observed gains are consistent and not due to random fluctuations.

Overall, these findings validate the advantages of the TSD method: (1) strong robustness across different levels of observation missingness and consistently outperforms both POP and Forecast-mae under all masking ratios; (2) consistent and statistically reliable performance improvement without sacrificing the base model’s accuracy; and (3) broad adaptability across different motion prediction architectures.

\noindent\textbf{Robustness with Continuous Masking.}~We conducted a comparative experiment on the robustness of the HiVT, HiVT-POP, and HiVT-TSD models under continuous masking conditions on the Argoverse 1 dataset, as shown in Table~\ref{tab:table2}. The experimental results indicate that, in partially observed scenarios, as the effective observation length decreases (from obs=20 to obs=1), the prediction performance of the original HiVT model exhibits a significant decline, a trend consistent across different hidden dimension settings (64 and 128). Although HiVT-POP shows some improvement over the baseline model at medium observation lengths (Obs= 5 to obs = 15), the enhancement is limited, and its performance under extremely short observation conditions (Obs=1) is even inferior to the baseline model, revealing the method's limitations in extreme scenarios. In contrast, our proposed TSD method performs excellently across all observation lengths, particularly outperforming HiVT-POP significantly under extremely short input lengths. These results fully validate the robustness of our method against continuous observation missingness and its adaptability to extremely short input lengths.

\noindent\textbf{Component Ablation.}~In Table~\ref{tab:comp_ab}, we analyze the gains of each component of the proposed TSD based on HiVT-TSD. The contents of the table show that by predicting accurate proxy target points and performing target-guided trajectory prediction, the model achieves improvements in both fully observed and partially observed trajectories, resulting in an overall performance enhancement. After adding self-distillation to the motion forecasting training, the model shows further improvement in fully observed trajectories and a significant increase in robustness for partially observed trajectory inputs.

\noindent\textbf{Self Distillation Ablation.}~We also conducted an ablation analysis of the self-distillation component in our proposed method. As shown in Table~\ref{tab:sd_ab}, by adding a branch for partially observed trajectory inputs and using the best mode from fully observed trajectory predictions as the gradient backpropagation mode for partial observations, the model can focus on the prediction results of the best mode under partial observations and perform backpropagation accordingly. After adding the MMD loss, the model further aligns the representations between fully observed and partially observed scenarios, thereby enhancing the model's robustness and overall performance.


\noindent\textbf{Qualitative Result.} To demonstrate the effectiveness of our method, we conducted qualitative experiments on the Argoverse1 dataset, as shown in \ref{fig:subfigures}. Case 1 shows a straight-driving scenario with an input history length of 20. Case 2 shows a turning scenario with an input history length of 20. It can be observed that our method, TSD, outperforms the original HIVT in both scenarios. In Case 1, the predicted trajectory of our method is closer to the ground truth than HIVT's prediction. In Case 2, HIVT incorrectly predicts a straight path, while our method successfully predicts the turning scenario. Case 3 shows an experiment where only part of the historical trajectory is provided, with a history length of 10. In this case, our method predicts the turning trajectory more closely to the ground truth compared to the original HIVT.

\begin{table}[!htbp]
  \centering
  \setlength\tabcolsep{2mm}
  \small
  \begin{tabular}{cc|ccc} 
    \toprule
    Model & Method & Training time & Memory &  Parameter \\ \midrule
    HiVT & origin & 32h & 6G & 0.65M \\
    HiVT & POP & 64h & 8G & 1.3M \\
    HiVT & TSD & 35h & 6G & 1.3M  \\
    DeMo & origin & 9h & 12G & 5.9M\\
    DeMo & POP & 20h & 18G & 11.8M \\
    DeMo & TSD & 12h & 12G & 6M \\
    \bottomrule
  \end{tabular}
  \caption{Computational cost compared to other methods.}
  \label{tab:cost}
\end{table}
\noindent\textbf{Computational cost compared to other method.}
We compared the computational cost of our method on different models in Table \ref{tab:cost}.Experiments on an NVIDIA RTX 4090 show that TSD provides a more favorable trade-off between efficiency and model complexity. Compared with baseline and POP variants, TSD consistently lowers both training time and memory usage while keeping parameter counts comparable. On DeMo, TSD reduces training time by 40\% and memory usage by 33\%, with similar improvements observed on HiVT. These results demonstrate that TSD offers a more efficient optimization strategy, achieving faster convergence and lower resource consumption without compromising model capacity.

\section{Conclution}

This paper presents a self-distillation motion forecasting method based on anchor-free target point guidance, tailored to the characteristics of partial observation trajectory prediction tasks. First, the method predicts long-term and short-term target points for different time horizons, which guide precise trajectory prediction under partial observation conditions. Second, we design a self-distillation approach for partial observation trajectories, ensuring that the feature distributions and prediction results remain consistent between full and partial observation scenarios. Extensive experiments demonstrate that the trained model achieves more accurate motion forecasting under partial observation and maintains, or even improves, performance under full observation.

  


{
    \small
    \bibliographystyle{ieeenat_fullname}
    \bibliography{main}
}


\end{document}